\documentclass{ecai} 

\usepackage{xcolor}
\usepackage{times}
\usepackage{colortbl}
\usepackage{latexsym}
\usepackage{amssymb}
\usepackage{amsmath}
\usepackage{amsthm}
\usepackage{booktabs}
\usepackage{enumitem}
\usepackage{graphicx}
\usepackage{color}
\usepackage{diagbox}
\usepackage{cleveref}
\usepackage{pifont}
\usepackage{makecell}
\usepackage{booktabs}
\usepackage{arydshln}
\usepackage{multirow}
\usepackage{subfigure}
\usepackage[graphicx]{realboxes}

\newcommand{\BibTeX}{B\kern-.05em{\sc i\kern-.025em b}\kern-.08em\TeX}

\begin{document}


\begin{frontmatter}




\title{Decomposing and Revising What Language Models Generate}


\author[A]{\fnms{Zhichao}~\snm{Yan}}
\author[B]{\fnms{Jiaoyan}~\snm{Chen}}
\author[C]{\fnms{Jiapu}~\snm{Wang}}
\author[D]{\fnms{Xiaoli}~\snm{Li}}
\author[A]{\fnms{Ru}~\snm{Li}$^{*,}$}
\author[E]{\fnms{Jeff Z.}~\snm{Pan}\thanks{Corresponding Authors}}

\address[A]{School of Computer and Information Technology, Shanxi University, Taiyuan, China}
\address[B]{Department of Computer Science, University of Manchester, Manchester, England}
\address[C]{Beijing University of Technology, Beijing, China}
\address[D]{Singapore University of Technology and Design, Singapore}
\address[E]{ILCC, School of Informatics, University of Edinburgh, Edinburgh, England}

\begin{abstract}
Attribution is crucial in question answering (QA) with Large Language Models (LLMs).
SOTA question decomposition-based approaches use long form answers to generate questions for retrieving related documents. However, the generated questions are often irrelevant and incomplete, resulting in a loss of facts in retrieval.
These approaches also fail to aggregate evidence snippets from different documents and paragraphs. 
To tackle these problems, we propose a new fact decomposition-based framework called FIDES (\textit{faithful context enhanced fact decomposition and evidence aggregation}) for attributed QA. 
FIDES uses a contextually enhanced two-stage faithful decomposition method to decompose long form answers into sub-facts, which are then used by a retriever
to retrieve related evidence snippets. If the retrieved evidence snippets conflict with the related sub-facts, such sub-facts will be revised accordingly.  
Finally, the evidence snippets are aggregated according to the original sentences. 
Extensive evaluation has been conducted with six datasets, with an additionally proposed new metric called $Attr_{auto-P}$ for evaluating the evidence precision. FIDES outperforms the SOTA methods by over 14\% in average with GPT-3.5-turbo, Gemini and Llama 70B series.
\end{abstract}

\end{frontmatter}


\section{Introduction}

\textit{Attributed Question Answering} (QA)~\cite{HCWQ+2025} aims to provide both an answer and an attribution report for a given question. Large Language Models (LLMs) have demonstrated impressive performance across various NLP tasks \cite{bohnet2022attributed, wang2024large}. However, LLMs often struggle with hallucination and indecisiveness~\cite{PRKS+2023,HVLP2025}, which undermine the reliability and trustworthiness. Thus, generating accurate attribution reports and correcting hallucinations are crucial for enhancing the practical utility of LLMs in real-world applications.

\begin{figure}[ht!]
  \centering 
  \includegraphics[height=9.5cm,width=8.5cm]{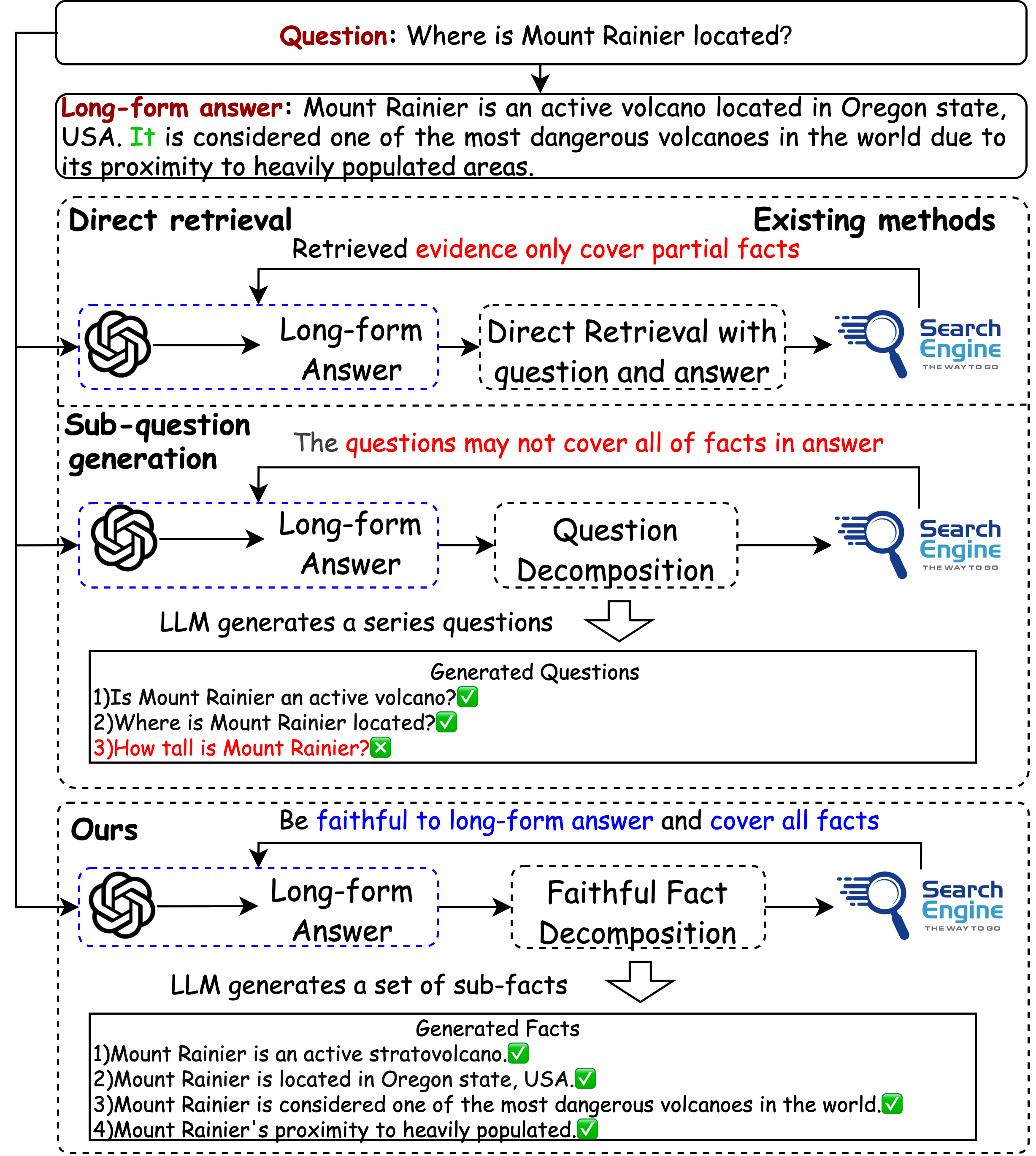}
  \caption{
  Existing methods typically cover only a portion of the facts in a long form answer during retrieval. Our approach, however, employs sub-facts of the answer for retrieval, ensuring comprehensive coverage of all the content in the answer.}
  \label{methods_compare}
  \end{figure}

The use of external resources for retrieval~\cite{SDVM+2025,HZWV+2024,WHHH+2024} is a mainstream solution for attributed QA. Depending on the timing of retrieval, these systems can be divided into two categories: Retrieval then Read  \citep{trivedi2022interleaving,muller-etal-2023-evaluating,gao2023enabling,li2023llatrieval}, and Post-hoc retrieval, which is challenging and still have many open problems, according to \citet{bohnet2022attributed}. 
It currently has two main paradigms: 1) Directly using the content generated by LLMs along with the question for retrieval \citep{bohnet2022attributed}, and 2) Question decomposition-based approaches that generate specific questions about the content for evidence retrieval ~\citep{gao2023rarr}.

As shown in Figure \ref{methods_compare}, the direct retrieval approach has a significant drawback: when using a long form answer to retrieve documents, it is challenging to cover all the content due to the multiple facts it contains. The question decomposition-based attribution method, which operates in two stages, both utilizing LLMs. In the first stage, retrieval, they prompt the LLM to generate a series of questions and use these questions to retrieve relevant documents with a retriever (such as a search engine). In the second stage, revising, the evidence is used to detect factual errors in the answer, and if any are found, the answer is revised based on the evidence.

A bottleneck of question decomposition-based method lies in the quality of the generated questions, which can sometimes be irrelevant to the given answers. In Figure \ref{methods_compare}, question decomposition-based method generates three questions, with the last question, ``\textit{How tall is Mount Rainier?}'', being irrelevant to the answer. An idealized question to cover this fact could be ``\textit{Why is Mount Rainier the most dangerous active volcano in the world?}''.
Furthermore, it fails to aggregate evidence snippets from different documents and paragraphs for sentence-level evaluation. For instance, the claim ``\textit{It is considered one of the most dangerous volcanoes in the world due to its proximity to heavily populated areas}'' encompasses two distinct facts and requires evidence snippets from different sources to support each.

\begin{figure*}[h]
  \centering 
  \includegraphics[height=9cm,width=17.6cm]{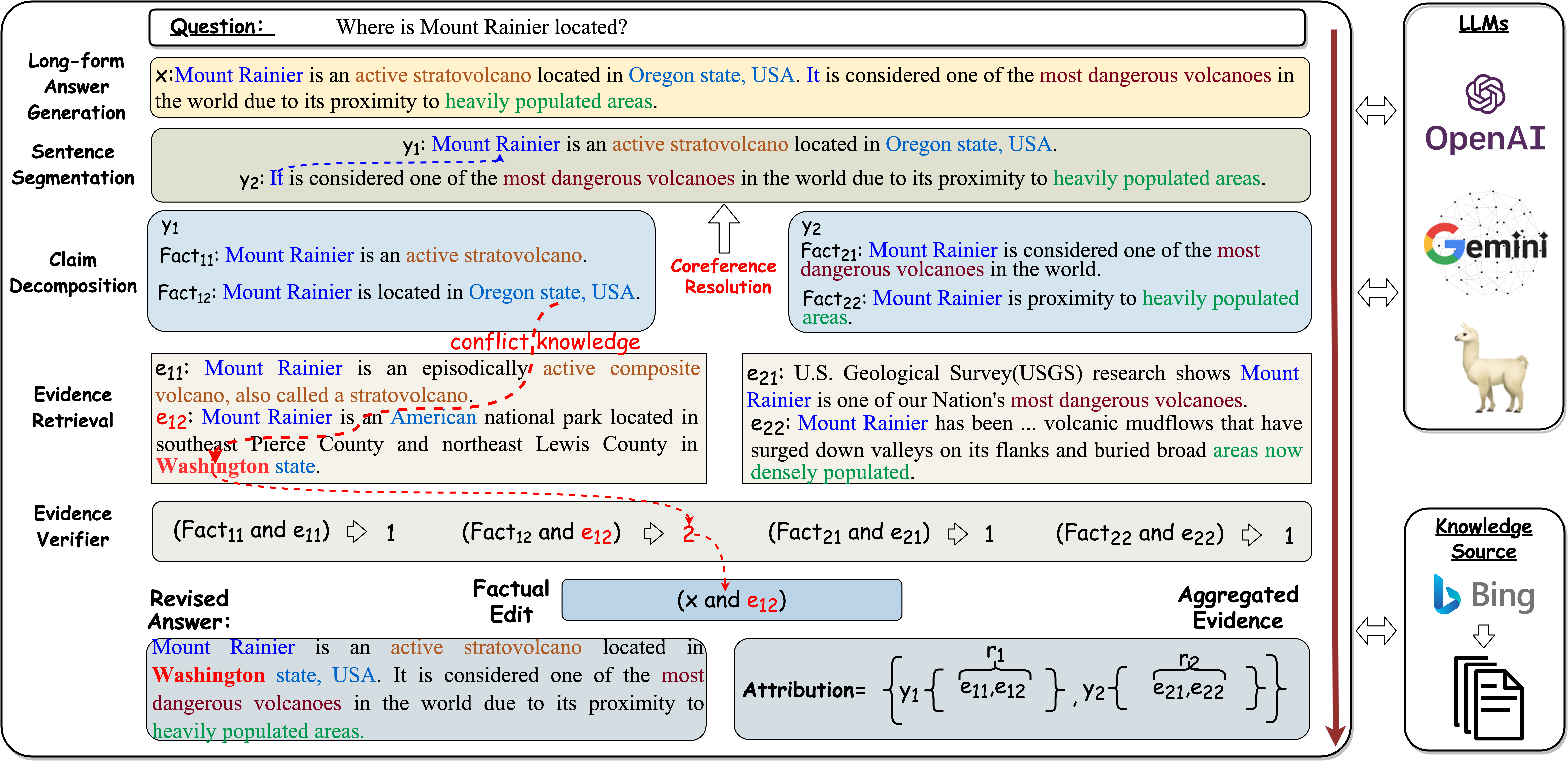}
  \caption{An overview of \textbf{FIDES}, which improves the attribution of the answer $x$ through fact decomposition and evidence aggregation. During
  the decomposition stage, $x$ is segmented into multiple facts, facilitating Evidence Retrieval (\textbf{ER}) from the Bing search. Evidence Verifier (\textbf{EV}) phase then evaluates 
  evidence against 
  facts, resulting in two states: \textbf{1} no conflicts detected, \textbf{2}, conflicting knowledge requiring Factual Edit
  (\textbf{FE}) of $x$ with the evidence. In the evidence aggregation stage, evidence is summarized based on the segmentation of $x$, generating an attribution report. 
  }
  \label{framework}
  \end{figure*}

Meanwhile, prior research has shown that fact decomposition is effective for retrieval and claim verification \citep{chen2023complex, lin-etal-2023-decomposing, wang2023hallucination}. It can faithfully represent the original answer and avoid generating irrelevant contents, ensuring the retrieval of all necessary evidence to support the answer.

To address the limitations of question decomposition-based method, we present a \textbf{F}a\textbf{I}thful context enhanced fact \textbf{DE}compo\textbf{S}ition and evidence aggregation (\textbf{FIDES}) strategy. The fact decomposition has two stages:  sentence segmentation (SS) which segments the answer into individual sentences, and claim decomposition (CD) which decomposes the sentence into sub-facts to retrieve evidence snippets. These snippets are then aggregated based on the SS results. In Figure \ref{methods_compare}, the answer is decomposed into four sub-facts after SS and CD. 
A key challenge of SS is coreference resolution (CR). For example, if the subject of the second sentence ``\textit{It is considered one of...}'', ``\textit{It}'', is not explicitly expressed as ``\textit{Mount Rainier}'', the sub-fact as the query is meaningless, and the search engine won't return relevant documents. So, 
we unify the CR and SS tasks, prompting the LLM to segment the answer and explicitly state the subject by a pronoun. Then, we follow RARR to construct the revise stage, the fact ``\textit{Mount Rainier...located in \textbf{Oregon} state, USA}'' will be revised into ``\textit{Mount Rainier...located in \textbf{Washington} state, USA}'' per the evidence $e_{12}$ in Figure \ref{framework}. Finally, the evidences from different documents and paragraphs will be aggregate for support the total sentence.

Moreover, in sentence-level attribution evaluation, the \textbf{$Attr_{auto}$} metric proposed by \cite{gao2023rarr} solely emphasizes evidence recall, leading to retrieval of numerous noisy documents and increased computational expenditure. To address this, we introduce a novel evaluation metric called $Attr_{auto-P}$, which accurately assesses the precision of retrieved evidence and mitigates the proportion of invalid retrieval.

Finally, our extensive experiments across Knowledge Graph~\cite{PVGW2017,PCEH+2017} Question Answering (KGQA), Open-domain QA, and multi-hop QA datasets reveal GPT-3.5-turbo as the top-performing fact decomposition model, Llama3 70B as the most effective model for editing, and FIDES as a significantly superior solution, surpassing state-of-the-art methods by over 14.2\% with GPT-3.5-turbo, Gemini, and Llama 70B series. Our contributions are summarized as follows:
\begin{itemize}
    \item This paper proposes a novel fact decomposition-based framework called FIDES (\textit{faithful context enhanced fact decomposition and evidence aggregation}) which integrated coreference resolution task with fact decomposition.
    \item This paper detailed analysis was conducted on the revision performance of different LLMs, and it was found that existing LLMs exhibit abnormal revisions in open modifications.
    \item This paper proposes a new metric to more comprehensively evaluation the attribution process.
\end{itemize}

\section{Related Work}
\textbf{Retrieval-based Attributed Question Answering}
From the perspective of retrieval timing, two trends have recently emerged: (1) Retrieval-then-Read (RTR) and (2) Post-hoc Retrieval. \citet{trivedi2022interleaving} introduced IRCoT, which interleaves retrieval with steps in a Chain of Thought (CoT), guiding the retrieval process with CoT and subsequently using the retrieved results to enhance the CoT. 
\citet{muller-etal-2023-evaluating} studied attribution for cross-lingual QA. \citet{gao2023enabling} introduced ALCE, which utilizes multiple methods for integrating retrieved documents into LLMs for answer generation. \citet{li2023llatrieval} proposed a progressive selection of evidence using LLMs with a classification-based prompted template. \citet{bohnet2022attributed} extensively evaluated LLM attributions, suggesting that the RTR approach achieves good performance but requires full use of a traditional training set, highlighting the potential of post-hoc retrieval. \citet{gao2023rarr} proposed RARR, the state-of-the-art method for post-hoc retrieval, with limitations discussed in Introduction. \citet{kang2023mitigating} proposed to combine RTR and post-hoc retrieval. Our approach addresses these limitations through fact decomposition and evidence aggregation, resulting in significant performance improvements. 

\textbf{Hallucination Detection and Revision}
Hallucination detection~\citep{chen2023hallucination, mishra2024fine, mndler2023selfcontradictory, wang2023hallucination} is a challenging task essential for enhancing the reliability of LLMs in real-world scenarios. In a similar vein, \citet{ZLLP2024} introduced TrustScore, the first effective evaluation metric for assessing the trustworthiness of LLM responses in a reference-free setting.
To address hallucinations, numerous model editing methods avoid updating the parameters of LLMs~\citep{gao2023rarr,HLTW+2023,song2024knowledge,wu2024updating,zheng-etal-2023-edit}. Unlike our approach, most of these methods focus solely on either hallucination detection or revision, not both.

\textbf{Fact Decomposition}
The technology of decomposition has been demonstrated that it can effectively solve 
complex questions, especially in various reasoning tasks~\citep{Ye_2023,zhang2023reasoning} and claim verification tasks~\citep{chen2023complex,lin-etal-2023-decomposing, wang2023hallucination}. Although the fact decomposition can faithfully generate sub-facts to represent the original answer, it fails to extract complete facts when pronouns appear. Particularly, if a fact includes a subject with a pronoun and is used as a query, the resulting sentence will fail to retrieve relevant documents.

\section{Approach}

Now we introduce the FIDES framework (cf. Figure \ref{framework}), which starts with a long form answer \textit{x} generated by a prompt and LLMs in response to a question. Initially, the framework prompts the model to divide \textit{x} into a series of short sentences $\{y_1,...,y_m\}$. Each sentence $y_m$ is then further broken down into a set of facts $\{Fact_{m1},...,Fact_{mk}\}$. The goal is for each $Fact_{mk}$ to encapsulate a single fact that can be used to retrieve relevant webpages from a search engine for evidence verification. If any conflicts arise during verification, an edit module will update \textit{x} using the gathered evidence. Finally, evidence snippets will be compiled as attributions $\{r_1,...,r_m\}$ for \textit{x}.

\subsection{Long form answer generation}
To better attribute the answer generated by a LLM 
we construct a long from answer following \citet{gao2023rarr}. To this end, we propose a robust method utilizes a new few-shot prompt template and different LLMs to generate a long form answer. 

\subsection{Fact decomposition (FD)}

We perform a two-state decoupling operation on the long form answer \textit{x}.In first stage, we segment $x$ into a set of sentences $\{y_1,y_2,...,y_m\}$, as illustrated in the \textbf{Sentence Segmentation (SS)} process in Figure \ref{framework}.
A straightforward method involves using heuristic rules, but it may result in sentences that omit critical entities, such as \textit{Mount Rainier} in the example, thereby failing to retrieve valuable webpages. Some existing works \citep{wang2023hallucination,gao2023enabling} prompt LLMs to decompose a document into sentences. However, this method heavily relies on the LLMs and cannot extract complete clauses when pronouns appear. Specifically, if a pronoun is used as the subject of the segmented sentence, the resulting sentence will miss relevant documents during retrieval.
To address this, we explicitly prompt LLMs to generate the subjects of sentences during segmentation, a process we call Explicit Coreference Resolution (ECR). 
Specifically, for the construction of the prompt template, we first provide the segmentation results of the long form answer, and then put the subject of original answer after it to unify modeling the sentence segment and coreference resolution tasks.

In second stage, \textbf{Claim Decomposition (CD)} decomposes a sentence $y_m$ into multiple atomic facts $\{Fact_{m1},...,Fact_{mk}\}$ by prompting a LLM. Ideally, the sub-fact only contains one fact, which to be as a query to retrieve evidence.
Both SS and CD were constructed using few-shot prompting, in addition, in order to extract the subject accurately, we add some instructions about the task based on few-shot prompt.

\begin{table*}[t!]
  
  \caption{\label{ChatGPT}
  Evaluation results on GPT-3.5-turbo-1106, Gemini, Llama2 70B and Llama3 70B. For sentence level attribution evaluation, we report both before and after editing (before$\to$after) \textbf{AR} and \textbf{AP}. For \textbf{AF1}, we report scores only after editing to clearly demonstrate the improvement achieved by FIDES. Additionally, we calculate the average AF1 score for each row as Average AF1 (\textbf{Ave-AF1}) to provide a comprehensive overview. Since the RARR paper does not provide access to the test datasets, we reproduce the results as reported in the publication.
  }
  \centering
  \resizebox{1\textwidth}{!}{
  \begin{tabular}{llllllllll:c}
    
  \cmidrule[\heavyrulewidth]{1-10}
  \multicolumn{1}{c}{\multirow{2}{*}{Method}} & \multicolumn{3}{c}{\textbf{WebQSP}}                             & \multicolumn{3}{c}{\textbf{Mintaka}}                                        & \multicolumn{3}{c}{\textbf{NQ}}                             \\ \cline{2-10} 
  \multicolumn{1}{c}{\rule{0pt}{8pt}}                        & AR$\uparrow$      & AP$\uparrow$  & AF1$\uparrow$    & AR$\uparrow$    & AP$\uparrow$ & AF1$\uparrow$    & AR$\uparrow$    & AP$\uparrow$ & \multicolumn{1}{l}{AF1 $\uparrow$} &     \\ \cmidrule(lr){1-10} \rowcolor{gray!40}
                                                & \multicolumn{8}{c}{\textbf{GPT-3.5-turbo-1106}}& \multicolumn{1}{l}{}                                                                                                                                                                    \\
  DRQA                                        & 0.588$\to$0.589 & 0.847$\to$0.847 & 0.694 & 0.428$\to$0.443 & 0.667$\to$0.680               & 0.537 & 0.423$\to$0.437 & 0.640$\to$0.660              & \multicolumn{1}{l}{0.527} &  \\
  DRA                                        & 0.609$\to$0.612 & \textbf{0.887}$\to$\textbf{0.887} & 0.724 & 0.436$\to$0.444 & 0.687$\to$0.693               & 0.541 & 0.437$\to$0.560 & 0.660$\to$0.667              & \multicolumn{1}{l}{0.530} & \\
  RARR                                        & 0.619$\to$0.645   & 0.625$\to$0.618 & 0.631 & 0.539$\to$0.624 & 0.514$\to$0.546              & 0.582 & 0.508$\to$0.556 & 0.503$\to$0.514                & \multicolumn{1}{l}{0.534} & \\
  FIDES                                        & \textbf{0.801}$\to$\textbf{0.799}   & 0.862$\to$0.862 & \textbf{0.829} & \textbf{0.671}$\to$\textbf{0.684} & \textbf{0.822}$\to$\textbf{0.836}               & \textbf{0.752}  & \textbf{0.665}$\to$\textbf{0.681}                & \textbf{0.756}$\to$\textbf{0.751}                              &  \multicolumn{1}{l}{\textbf{0.714}} &  \\ \rowcolor{gray!40}
                                              & \multicolumn{8}{c}{\textbf{Gemini}}&\multicolumn{1}{l}{}                                                                                                                                                                     \\
  DRQA                                        & 0.401$\to$0.401 & 0.892$\to$0.892 & 0.553 & 0.404$\to$0.403 & 0.787$\to$0.778               & 0.531 & 0.335$\to$0.349 & 0.764$\to$0.791              & \multicolumn{1}{l}{0.484} & \\
  DRA                                        & 0.410$\to$0.410 & \textbf{0.899}$\to$\textbf{0.899} & 0.563 & 0.398$\to$0.397 & 0.873$\to$0.860               & 0.541 & 0.353$\to$0.367 & 0.810$\to$0.824               & \multicolumn{1}{l}{0.508} & \\
  RARR                                        & 0.559$\to$0.603 & 0.589$\to$0.656 & 0.605 & 0.489$\to$0.520 & 0.501$\to$0.527               & 0.523 & 0.454$\to$0.553 & 0.493$\to$0.557               & \multicolumn{1}{l}{0.555} & \\
  FIDES                                        & \textbf{0.732}$\to$\textbf{0.733} & 0.859$\to$0.857 & \textbf{0.790} & \textbf{0.717}$\to$\textbf{0.718}                & \textbf{0.823}$\to$\textbf{0.831}                               &\textbf{0.770}                  &\textbf{0.633}$\to$\textbf{0.651}                 &\textbf{0.807}$\to$\textbf{0.820}                              &\multicolumn{1}{l}{\textbf{0.726}}  &  \\ \rowcolor{gray!40}
  & \multicolumn{8}{c}{\textbf{Llama2 70B}}    &\multicolumn{1}{l}{}                                                                                                                                                                 \\
  DRQA                                        & 0.521$\to$0.521 & 0.765$\to$0.765 & 0.620 & 0.365$\to$0.372 & 0.593$\to$0.600               & 0.460 & 0.408$\to$0.423 & 0.601$\to$0.615               & \multicolumn{1}{l}{0.501} & \\
  DRA                                        & 0.531$\to$0.531 & 0.758$\to$0.758 & 0.625 & 0.374$\to$0.384 & 0.593$\to$0.600               & 0.468 & 0.418$\to$0.418 & 0.635$\to$0.635               & \multicolumn{1}{l}{0.504}  & \\
  RARR                                        & 0.629$\to$0.628 & 0.399$\to$0.541 & 0.582 & 0.443$\to$0.526 & 0.312$\to$0.437               & 0.477 & \textbf{0.443}$\to$\textbf{0.540} & 0.297$\to$0.434               & \multicolumn{1}{l}{0.481} & \\
  FIDES                                        & \textbf{0.699}$\to$\textbf{0.699} & \textbf{0.844}$\to$\textbf{0.844} & \textbf{0.765} & \textbf{0.585}$\to$\textbf{0.595}                & \textbf{0.762}$\to$\textbf{0.782}                               &\textbf{0.676}                  &0.531$\to$0.532                 &\textbf{0.674}$\to$\textbf{0.671}                              &\multicolumn{1}{l}{\textbf{0.593}} &  \\  \rowcolor{gray!40}
  & \multicolumn{8}{c}{\textbf{Llama3 70B}}   & \multicolumn{1}{l}{}                                                                                                                                                                 \\
  DRQA                                        & 0.435$\to$0.435 & 0.747$\to$0.747 & 0.550 & 0.377$\to$0.377 & 0.733$\to$0.733               & 0.498 & 0.378$\to$0.378 & 0.707$\to$0.707               & \multicolumn{1}{l}{0.492} & \\
  DRA                                        & 0.435$\to$0.435 & 0.760$\to$0.760 & 0.553 & 0.388$\to$0.388 & 0.733$\to$0.733               & 0.507 & 0.384$\to$0.384 & 0.713$\to$0.713               & \multicolumn{1}{l}{0.499} & \\
  RARR                                        & 0.605$\to$0.624 & 0.520$\to$0.511 & 0.562 & 0.529$\to$0.593 & 0.501$\to$0.502               & 0.544 & \textbf{0.515}$\to$\textbf{0.633} & 0.506$\to$0.520               & \multicolumn{1}{l}{0.571} & \\
  FIDES                                        & \textbf{0.739}$\to$\textbf{0.755} & \textbf{0.816}$\to$\textbf{0.831} & \textbf{0.791} & \textbf{0.675}$\to$\textbf{0.682}                & \textbf{0.751}$\to$\textbf{0.751}                               &\textbf{0.715}                  &0.633$\to$0.645                 &\textbf{0.706}$\to$\textbf{0.713}                              &\multicolumn{1}{l}{\textbf{0.678}} & \\ \midrule
  \multicolumn{1}{c}{} & \multicolumn{3}{c}{\rule{0pt}{8pt}\textbf{StrategyQA}}                             & \multicolumn{3}{c}{\textbf{HotpotQA}}                                        & \multicolumn{3}{c}{\textbf{Musique}}                             \\ \cline{1-11} 
  \multicolumn{1}{c}{\rule{0pt}{8pt}}                        & AR$\uparrow$      & AP$\uparrow$  & AF1$\uparrow$    & AR$\uparrow$    & AP$\uparrow$ & AF1$\uparrow$    & AR$\uparrow$    & AP$\uparrow$ & \multicolumn{1}{l}{AF1 $\uparrow$}  & Ave-AF1$\uparrow$  \\ \midrule
  \rowcolor{gray!40}
  & \multicolumn{9}{c}{\rule{0pt}{8pt}\textbf{GPT-3.5-turbo-1106}} &                                                                                                                                                                    \\
  DRQA                                        & 0.237$\to$0.241 & 0.490$\to$0.490 & 0.323 & 0.285$\to$0.297 & 0.526$\to$0.545               & 0.384 & - & -              & - & 0.470 \\
  DRA                                        & 0.258$\to$0.260 & \textbf{0.557}$\to$\textbf{0.584} & 0.367 & 0.305$\to$0.315 & 0.520$\to$0.531               & 0.395 & - & -              & - & 0.490\\
  RARR                                        & 0.281$\to$0.325   & 0.272$\to$0.292 & 0.308 & 0.384$\to$0.519 & 0.329$\to$0.422              & 0.466 & - & -                & - & 0.485\\
  FIDES                                        & \textbf{0.479}$\to$\textbf{0.490}   & 0.581$\to$0.583 & \textbf{0.527} & \textbf{0.608}$\to$\textbf{0.617} & \textbf{0.664}$\to$\textbf{0.675}               & \textbf{0.644}  & -                & -                              &  - & \textbf{0.676} \\ \rowcolor{gray!40}
                                              & \multicolumn{9}{c}{\textbf{Gemini}} &                                                                                                                                                                     \\
  DRQA                                        & 0.347$\to$0.347 & 0.780$\to$0.780 & 0.480 & 0.299$\to$0.299 & 0.620$\to$0.620               & 0.403 & 0.251$\to$0.251 & 0.600$\to$0.600               & 0.354 & 0.468\\
  DRA                                        & 0.349$\to$0.352 & \textbf{0.800}$\to$\textbf{0.800} & 0.489 & 0.294$\to$0.312 & 0.613$\to$0.647               & 0.421 & 0.262$\to$0.274 & 0.630$\to$0.650               & 0.385 & 0.485\\
  RARR                                        & 0.525$\to$0.564 & 0.517$\to$0.537 & 0.549 & 0.497$\to$0.656 & 0.477$\to$0.545               & 0.595 & 0.471$\to$0.591 & 0.400$\to$0.441               & 0.505 & 0.556\\
  FIDES                                        & \textbf{0.622}$\to$\textbf{0.629} & 0.745$\to$0.751 & \textbf{0.684} & \textbf{0.566}$\to$\textbf{0.587}                & \textbf{0.609}$\to$\textbf{0.705}                               &\textbf{0.641}                  &\textbf{0.521}$\to$\textbf{0.526}                 &\textbf{0.611}$\to$\textbf{0.609}                              &\textbf{0.564} & \textbf{0.696}\\ \rowcolor{gray!40}
  & \multicolumn{9}{c}{\textbf{Llama2 70B}} &                                                                                                                                                                    \\
  DRQA                                        & 0.271$\to$0.276 & 0.467$\to$0.487 & 0.351 & 0.208$\to$0.208 & 0.333$\to$0.333               & 0.255 & 0.189$\to$0.189 & 0.353$\to$0.353               & 0.246 & 0.406\\
  DRA                                        & 0.312$\to$0.312 & 0.520$\to$0.520 & 0.390 & 0.280$\to$0.306 & 0.607$\to$0.660               & 0.418 & 0.198$\to$0.293 & 0.367$\to$0.380               & 0.265 & 0.445\\
  RARR                                        & \textbf{0.418}$\to$\textbf{0.540} & 0.274$\to$0.422 & 0.473 & 0.443$\to$0.601 & 0.477$\to$0.512               & 0.553 & \textbf{0.335}$\to$\textbf{0.523} & 0.222$\to$0.374               & 0.436 & 0.500\\
  FIDES                                        & 0.502$\to$0.507 & \textbf{0.666}$\to$\textbf{0.679} & \textbf{0.580} & \textbf{0.587}$\to$\textbf{0.626}                & \textbf{0.677}$\to$\textbf{0.685}                               &\textbf{0.654}                  &0.402$\to$0.409                 &\textbf{0.592}$\to$\textbf{0.595}                              &\textbf{0.484} & \textbf{0.630}
  \\ \rowcolor{gray!40}
  & \multicolumn{9}{c}{\textbf{Llama3 70B}}  &                                                                                                                                                                     \\
  DRQA                                        & 0.236$\to$0.236 & 0.627$\to$0.627 & 0.343 & 0.259$\to$0.259 & 0.553$\to$0.553               & 0.353 & 0.216$\to$0.216 & 0.523$\to$0.523               & 0.305 & 0.424\\
  DRA                                        & 0.237$\to$0.237 & 0.613$\to$0.613 & 0.342 & 0.262$\to$0.262 & 0.560$\to$0.560               & 0.357 & 0.215$\to$0.215 & 0.510$\to$0.510               & 0.303 & 0.427\\
  RARR                                        & 0.394$\to$0.479 & 0.369$\to$0.396 & 0.433 & 0.412$\to$0.608 & 0.393$\to$0.423               & 0.499 & \textbf{0.367}$\to$\textbf{0.507} & 0.366$\to$0.316               & 0.390 & 0.500\\
  FIDES                                        & \textbf{0.497}$\to$\textbf{0.505} & \textbf{0.582}$\to$\textbf{0.589} & \textbf{0.544} & \textbf{0.512}$\to$\textbf{0.537}                & \textbf{0.586}$\to$\textbf{0.599}                               &\textbf{0.566}                  &0.461$\to$0.479                 &\textbf{0.560}$\to$\textbf{0.563}                              &\textbf{0.518}  & \textbf{0.636}          \\ \bottomrule
              \end{tabular}
  
  }
     
  \end{table*}

\subsection{Evidence retrieval and verification}

\textbf{Evidence Retrieval (ER)}. Each fact $Fact_{mk}$ is used to retrieve the top $M=5$ relevant webpages; then, we utilized a sliding window to select the relevant context chunk. We maintained the four sentences in every chunk and used the cross-encoder model \citep{reimers-2019-sentence-bert}, which trained on MS MARCO \citep{nguyen2016ms} to rerank the chunks. In the last, we select top $J=1$ text as the evidence of $Fack_{mk}$.

Given the potential for LLMs to generate output containing conflicts or illusory knowledge compared to retrieved evidence, we employ an \textbf{Evidence Verifier (EV)} to determine whether $x$ needs revision. The EV can output two states: \textbf{1} which indicates that according to the retrieved evidence, in $Fact_{mk}$, no knowledge conflicts has been detected, \textbf{2} which indicates that there are conflicted knowledge between the evidence and $Fact_{mk}$. EV is constructed by few-shot prompting.

\subsection{Factual edit and evidence aggregate}
\noindent \textbf{Factual Edit (FE)}.
When the \textbf{EV} model outputs state \textbf{2} indicating that $x$ has conflicts with the retrieved knowledge from the search engine, $x$ will be edited. To this end, we construct a prompt template using few-shot prompting, with each example in the template implemented using Chain-of-Thought (CoT) prompting \citep{NEURIPS2022_9d560961}. 

\noindent \textbf{Evidence Aggregation (EA)}. 
For $y_m$, which consists of ${Fact_{m1}, Fact_{m2}, \ldots, Fact_{mk}}$, the corresponding pieces of evidence ${e_{m1}, e_{m2}, \ldots, e_{mk}}$ are aggregated into a sequence named $r_m$. Since $Fact_k$ is obtained through the decoupling of $y_m$, the pieces of evidence $e_{mk}$ may overlap. Therefore, we eliminate duplicated snippets. Finally, the Attribution Report (ARt) is defined as $ARt = {[y_1, r_1], [y_2, r_2], \ldots, [y_m, r_m]}$.

\section{Experiments}
This section mainly introduces the evaluation setups, experimental results and some analysis.

\subsection{Evaluation Setups}

\textbf{Dataset}.
We perform extensive experiments on six datasets -- Natural Questions (NQ) \citep{kwiatkowskinatural} for open-domain QA, WebQSP \citep{yih-etal-2016-value} and Mintaka \citep{sen2022mintaka} for KGQA, and three Multihop datasets: StrategyQA \citep{geva2021strategyqa}, HotpotQA \citep{yang2018hotpotqa} and Musique \citep{Trivedi2021MM}. Following \citep{gao2023rarr}, we randomly select 150 samples from each of the test datasets.

\noindent \textbf{Attribution Corpus}.
We use Microsoft Bing\footnote{https://api.bing.microsoft.com/v7.0/search/} for document evidence retrieval.

\noindent \textbf{Baselines}.
We compare FIDES with three post-hoc retrieval enhancement systems.  The first two are Direct Retrieval with Question and Answer (DRQA) \citep{bohnet2022attributed} and RARR \citep{gao2023rarr}. We also implement another baseline model based on the DRQA, which Directly Retrieves documents by the original long form Answer (\textbf{DRA}). It can be used to demonstrate which LLM gets the best performance in the long form answer generation module without any decomposition.

\noindent \textbf{LLMs}.
We utilize GPT-3.5-turbo-1106, Gemini 1.0, Llama2 70B and Llama3 70B as the backbone model.

\noindent \textbf{Metrics}.
\citet{gao2023rarr} proposed a sentence level attribution score $Attr_{auto}$. For each sentence \textit{y} of the answer \textit{x}, and for each evidence snippet \textit{r} in \textit{A}, where \textit{A} = [$r_1,r_2,...,r_j,..,r_m$]. Let $NLI(r,y)$ be the model probability of \textit{r} entailing \textit{y}. $NLI(\bullet)$ is based on a model called TRUE \footnote{https://huggingface.co/google/t5\_xxl\_true\_nli\_mixture} , which implemented by a T5 checkpoint with 11B parameters and fine-tuned on a collection of natural language inference datasets \citep{honovich-etal-2022-true}. This model uses ``premise: \{PREMISE\} hypothesis: \{\}'' as prompt template. 
This metric is expressed as:
\begin{equation}
  Attr_{auto}(x,A)= \underset{y \in x}{avg} \  \underset{r \in A}{max} NLI(r,y). \label{attr}
\end{equation}

However, $Attr_{auto}$ will always get a high score if the number of evidence snippets is abundant enough, i.e., $Attr_{auto}$ solely considers the recall, but neglects the precision. In light of this, we propose $Attr_{auto-P}$ to assess the precision of retrieved information. Based on the \textit{NLI} model, we define
\begin{equation}
\hspace{-0.7cm}
\begin{aligned}
  &Attr_{auto-P}(x,A) = \\
  &\frac{\sum_{j=1}^{m}\left
   \|r_j\!\mid 
    \!(\sum_{i=1}^{m}
    \mathbb{I}
    (NLI(r_j,y_i) \!=\!1 ) / m)
     \!=\!1 \right \| }
    {\left
     \| A 
     \right \| 
    }  \label{attrn}
\end{aligned}
\end{equation}
for each $r \in A$, it undergoes comparison with every $y \in x$ where $m$ is the number of sentences in $x$. If all the $NLI(r_j,y_i)$ scores are not equal to 1, $r_j$ is defined as a noisy snippet. Here, $\| \cdot  \| $ counts the number of tokens. $\mathbb{I}$ signifies whether the $r$ entails $y$; it returns \textbf{1} for True and \textbf{0} for False. The $(\sum_{i=1}^{m}\mathbb{I}(NLI(r_j,y_i)=1)/m=1$ denotes $r_j$ for every $y_i$ is useless.
To distinguish $Attr_{auto-P}$ from $Attr_{auto}$ that focuses on recall, we denote $Attr_{auto}$ as $Attr_{auto-R}$.
To assess the validity of $Attr_{auto-P}$, we randomly selected 50 samples for human evaluation in each test dataset
(see section \ref{sec:eval}).

We denote $Attr_{auto-P}$ as \textbf{AP} and $Attr_{auto-R}$ as \textbf{AR} for simplicity. We also calculate  $Attr_{auto-F1}$ (\textbf{AF1}) as 2*AP*AR/(AP+AR), taking both precision and recall into consideration.

\subsection{Experimental results}
From Table \ref{ChatGPT}, we observe that FIDES demonstrates significant improvements in sentence-level attribution evaluation metrics across six datasets on all LLMs. The improvements in Average AF1 score (Ave-AF1) are \textbf{0.191}, \textbf{0.14}, \textbf{0.13}, and \textbf{0.136} across two KGQA datasets, one open-domain QA dataset and three multi-hop reasoning datasets with GPT-3.5-turbo, Gemini, Llama2, and Llama3, respectively, demonstrating FIDES' superior effectiveness and generalization capabilities.

On both KGQA and open-domain QA datasets, GPT-3.5, Gemini, and Llama3 achieve similar results, with Llama2 performing slightly worse. For StrategyQA, HotpotQA, and Musique, the Gemini-based method yields the best performance, indicating that Gemini excels in multi-hop attributed question-answer tasks. Due to GPT-3.5's knowledge base being limited to 2021, many questions in Musique cannot be answered. Consequently, we only use Gemini, Llama2, and Llama3 for evaluation on this dataset.

We observe that the baseline model DRA sometimes achieves higher AP scores than RARR and FIDES, but lower AR scores. This is because AP evaluates the precision of the retrieved evidence. DRA directly retrieves documents using a long form answer, resulting in most returned documents being valid for supporting the answers. However, it fails to cover all the facts in the long form answer, leading to higher AP scores but lower AR scores. By decoupling the long form answer and retrieving corresponding documents for sub-facts, FIDES achieves both high AP scores and a high coverage rate of the facts in the long form answer, resulting in the best overall performance.
Furthermore, comparing the results of DRA built using GPT-3.5, Gemini, and the Llama series across six datasets, GPT-3.5 demonstrates the best performance as an answer generation module on WebQSP, Mintaka, and NQ. Meanwhile, Gemini achieves the best performance on three multi-hop datasets.

Moreover, we observe that the AR or AP scores may remain unchanged or even decrease after editing. This phenomenon suggests that the edit module does not always perform effective edits, which may be due to two main factors: 1) the \textbf{EV} model may fail to detect some fine-grained errors, and 2) the \textbf{FE} module may introduce incorrect edits. Although this is not the expected outcome, it warrants further investigation.

\begin{table}[t]
\caption{\label{ablation}
\textbf{Ablation results} of FIDES. We report the results after editing on GPT-3.5-turbo. ``w/o SS" denotes 
removing sentence segmentation but directly applying claim decomposition towards the answer $x$ for evidence retrieval. ``w/o CD" indicates removing claim decomposition but directly using the segmented sentences for evidence retrieval ``w/o CD \& w/ QD" involves replacing claim decomposition by question decomposition (QD), keeping evidence aggregation. ``w/o ECR" depicts removing the explicit conference resolution prompt in SS.}
\centering

\setlength{\tabcolsep}{0.55cm}{
\begin{tabular}{lccc}
\toprule
\multirow{2}{*}{Method}      & \multicolumn{1}{l}{AR$\uparrow$} & \multicolumn{1}{l}{AP$\uparrow$} & \multicolumn{1}{l}{AF1$\uparrow$} \\ \cline{2-4}
      & \multicolumn{3}{c}{\textbf{WebQSP}}           \\ \hline
FIDES  & \textbf{0.799}       & \textbf{0.862}      & \textbf{0.829}     \\
w/o SS & 0.741       & 0.713      & 0.727     \\ 
w/o CD \& w/ QD & 0.703       & 0.818      & 0.756     \\
w/o CD & 0.740       & 0.826      & 0.780     \\ 
w/o ECR & 0.782       & 0.817      & 0.779     \\\hline
      & \multicolumn{3}{c}{\textbf{Mintaka}}          \\ \hline
FIDES  & \textbf{0.684}       & \textbf{0.836}      & \textbf{0.752}     \\
w/o SS & 0.611       & 0.622      & 0.617     \\ 
w/o CD \& w/ QD & 0.607       & 0.723      & 0.660 \\ 
w/o CD & 0.591       & 0.699      & 0.641     \\
w/o ECR & 0.683       & 0.759      & 0.709     \\\hline
      & \multicolumn{3}{c}{\textbf{NQ}} \\ \hline
FIDES  & \textbf{0.681}       & \textbf{0.751}       & \textbf{0.714}     \\
w/o SS & 0.627       & 0.580      & 0.603     \\ 
w/o CD \& w/ QD & 0.564       & 0.646      & 0.602 \\ 
w/o CD & 0.597       & 0.645      & 0.620 \\
w/o ECR & 0.666       & 0.714      & 0.669 \\ \hline
      & \multicolumn{3}{c}{\textbf{StrategyQA}} \\ \hline
FIDES  & \textbf{0.490}       & \textbf{0.583}       & \textbf{0.527}     \\
w/o SS & 0.380       & 0.390      & 0.385     \\ 
w/o CD \& w/ QD &   0.407     &    0.499   & 0.448  \\ 
w/o CD &  0.375      &  0.408     &0.391  \\
w/o ECR & 0.482       & 0.550      & 0.484 \\\hline
      & \multicolumn{3}{c}{\textbf{HotpotQA}} \\ \hline
FIDES  & \textbf{0.617}       & \textbf{0.675}       & \textbf{0.645}     \\
w/o SS & 0.503       & 0.510      & 0.496     \\ 
w/o CD \& w/ QD &  0.431      &   0.453    & 0.441 \\ 
w/o CD & 0.454       & 0.479      & 0.466\\     
w/o ECR & 0.602       & 0.664      & 0.611 
\\\bottomrule
\end{tabular}
}

\end{table}

\subsection{Ablation study}

Table \ref{ablation} presents the main ablation study results of FIDES when different modules are removed or replaced. Our observations are as follows:
(1) Both SS and CD play a positive role in FIDES, as removing  either module results in a noticeable drop in performance.
(2) Comparing w/o SS and w/o CD reveals that the SS module plays a more pivotal role than the CD module. This can be attributed to the reliance on CR in the SS module, as its absence will make CD fail to retrieve useful evidence from the search engine and multi evidence snippets aggregation based on SS results.
(3) Comparing w/o CD with w/o CD \& w/ QD indicates claim decomposition is also positive besides evidence aggregation.  
(4) The results of w/o ECR, particularly the decline 5\% in AF1 in WebQSP and the 4.3\% score in NQ, underscore the importance of explicit context resolution (ECR) in SS for effective performance.

\begin{figure}[t]
\centering
\subfigure[ AF1 scores on Mintaka before editing]{\includegraphics[height=3cm,width=4.1cm]{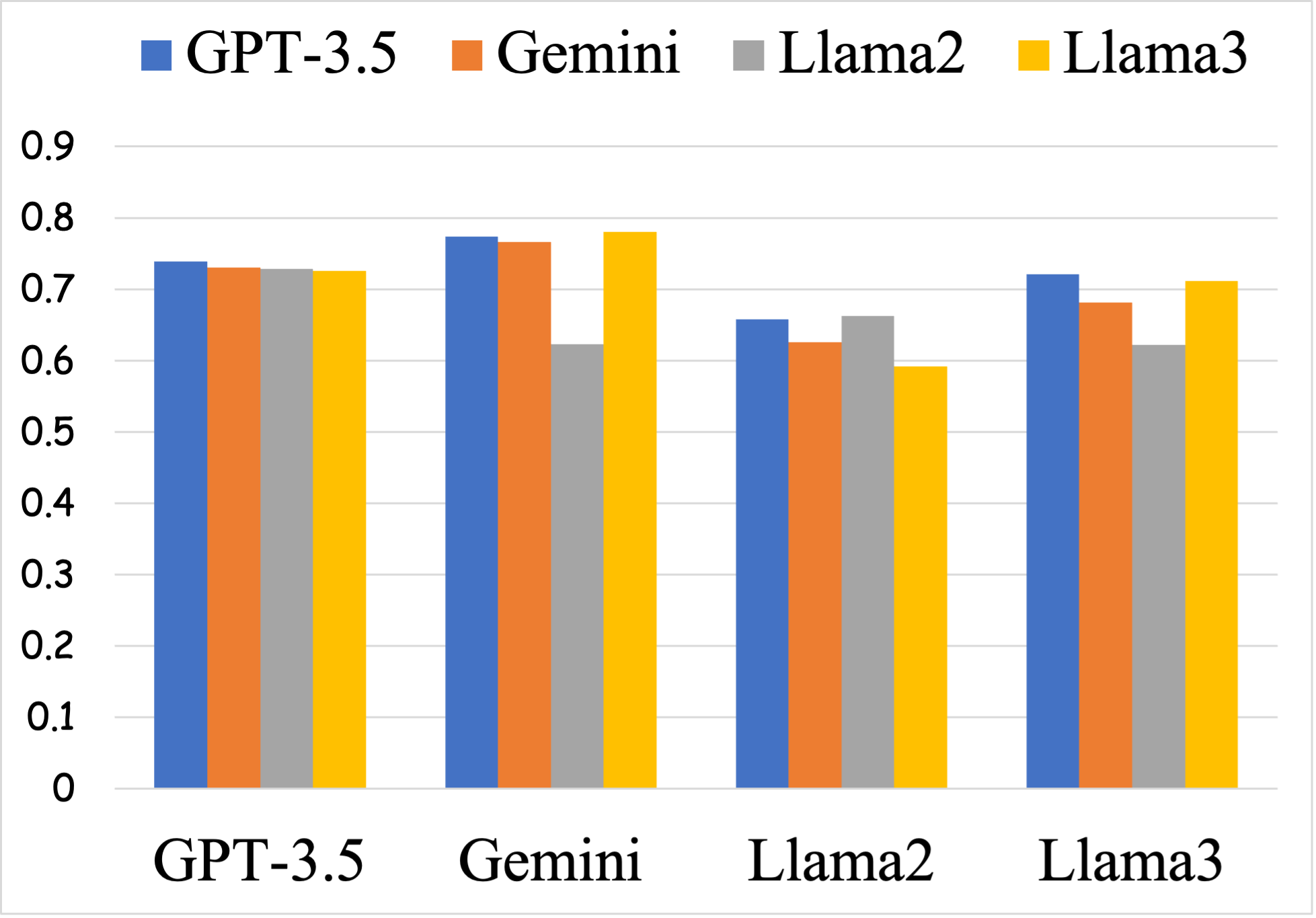}\label{Mintaka_count}}
\subfigure[ AF1 scores on NQ before editing]{\includegraphics[height=3cm,width=4.1cm]{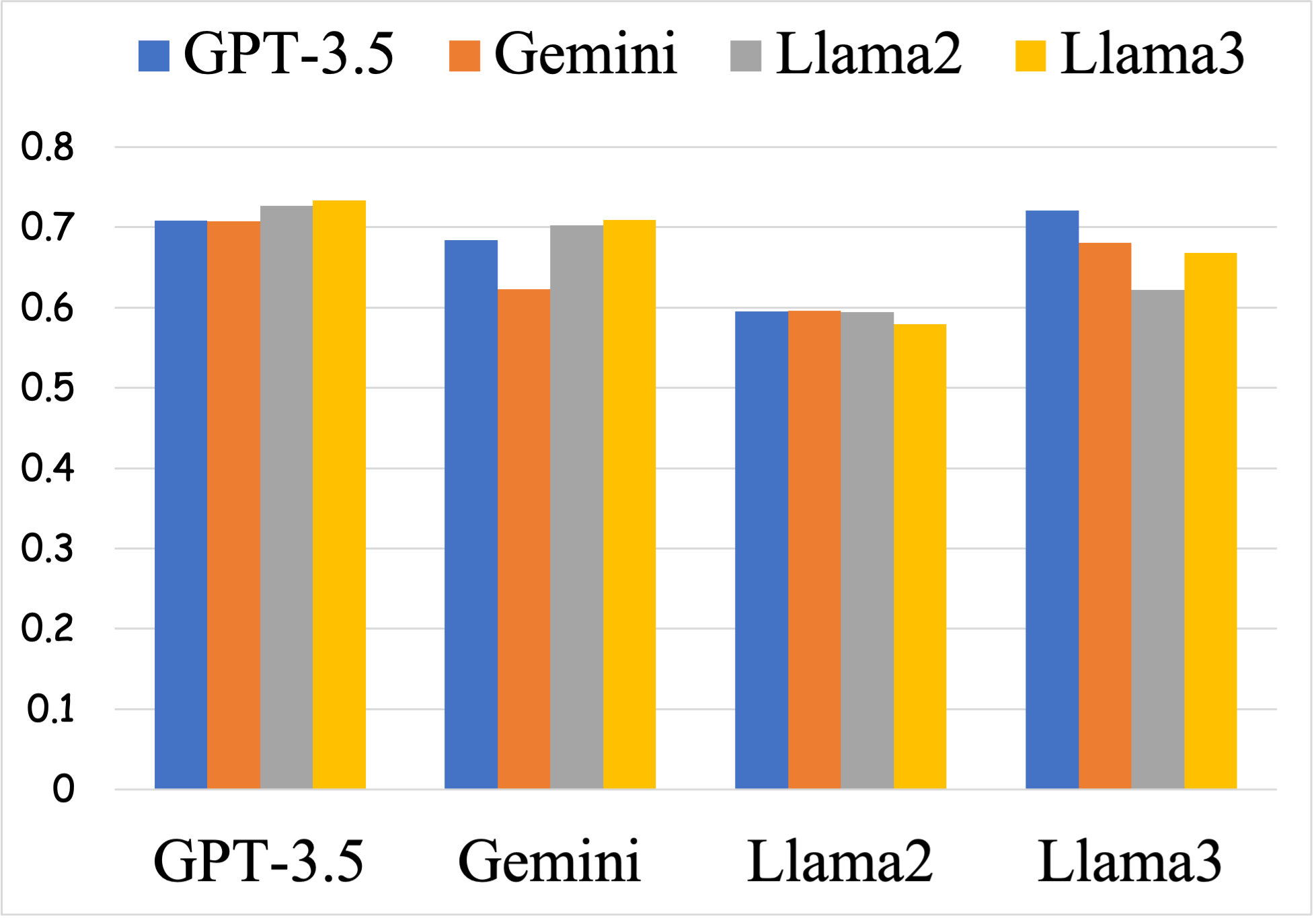}\label{NQ_count}}
\subfigure[ AF1 scores on Musique before editing]{\includegraphics[height=3cm,width=4.1cm]{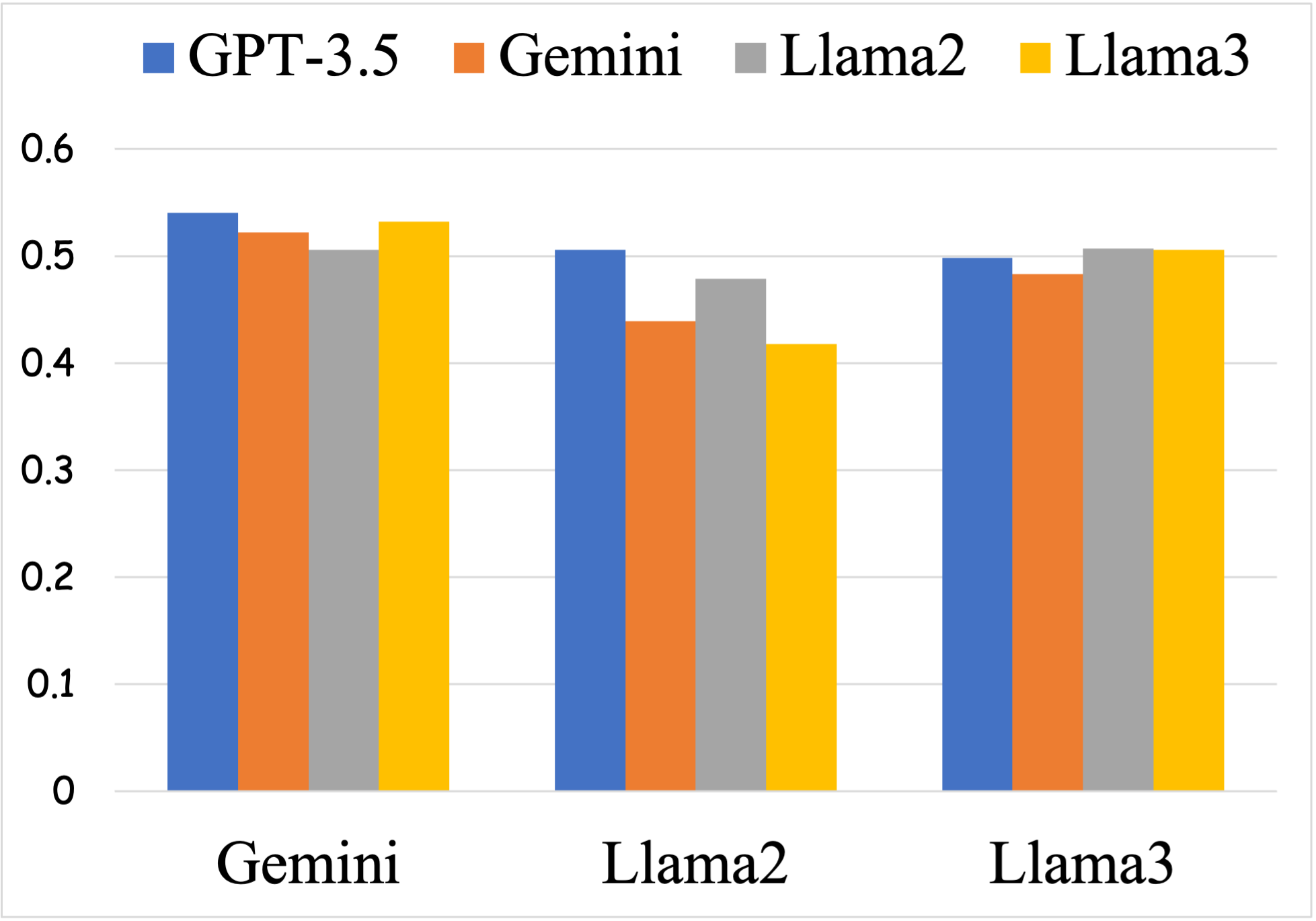}\label{musique_count}}
\subfigure[ The times of each LLM achieves the best performance when used it for FD in experiments.]{\includegraphics[height=3cm,width=4.1cm]{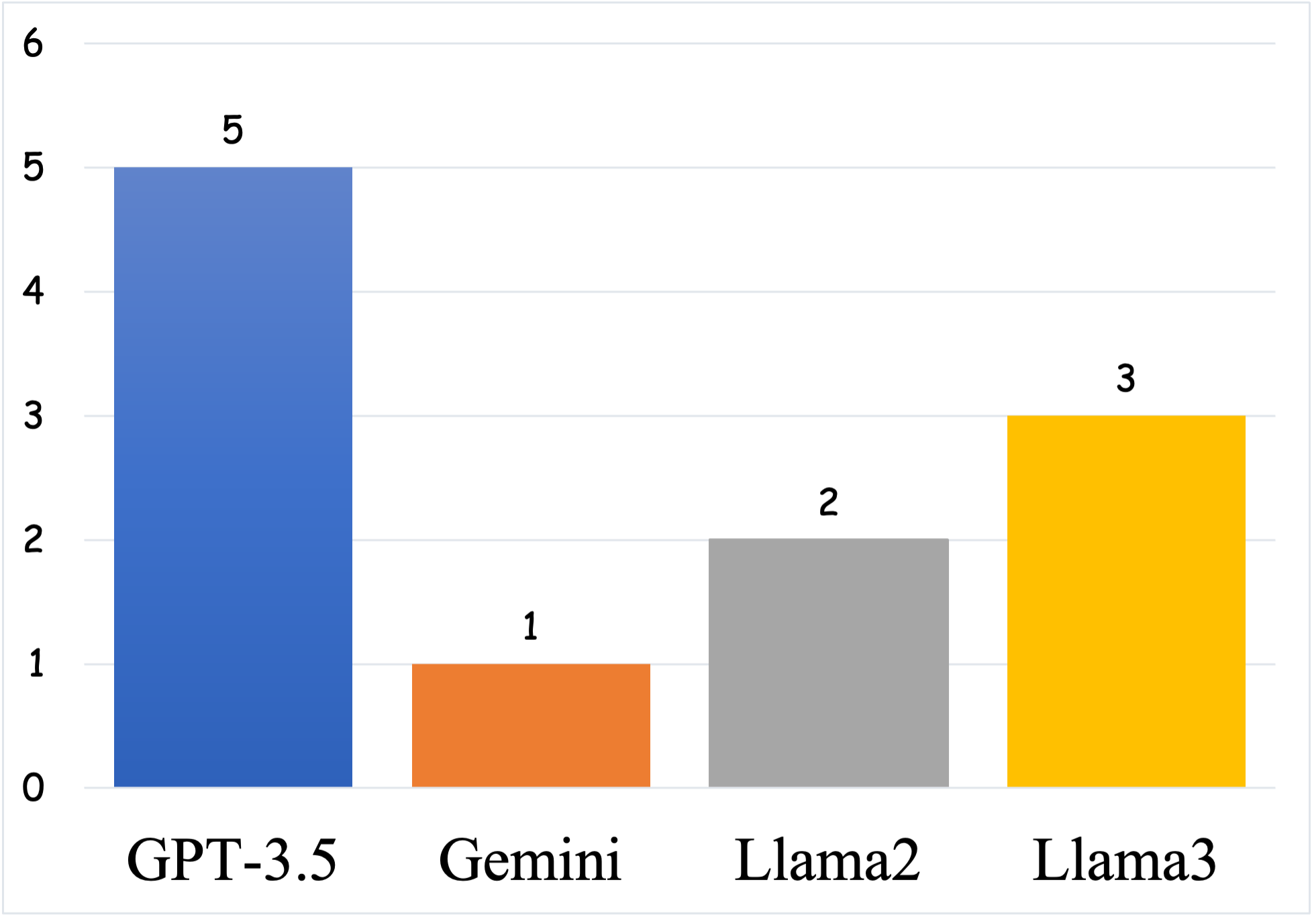}\label{total_count}}
\caption{ Figures (a), (b) and (c) show the experimental results (averaged AF1 of FIDES without FE and EV) of using different LLMs as the model for long form answer generation and the model for FD (i.e., SS and CD) on Mintaka, NQ, and Musique before editing. The x-axis represents the use of GPT-3.5, Gemini, Llama2, and Llama3 for long form answer generation respectively, while the legend indicates the use of these four LLMs for FD. }
\label{fig4}
\end{figure}

\begin{table}[ht]
\centering
\setlength{\tabcolsep}{0.15cm}{
\begin{tabular}{cccccc}
\toprule
\multicolumn{4}{c}{Modules} & \multicolumn{2}{c}{Metrics} \\ \hline
\textbf{AG} & \textbf{FD} & \textbf{EV} & \textbf{FE} & AR$\uparrow$ & AP$\uparrow$ \\ \midrule
\multirow{8}{*}{GPT3.5} & \multirow{3}{*}{Gemini} & Gemini & Gemini & 0.662$\to$0.665 & 0.834$\to$0.841 \\
&& Llama2 & Gemini & 0.643$\to$0.649 & 0.817$\to$0.823 \\
&& Gemini & Llama2 & 0.657$\to$0.661 & 0.817$\to$0.810 \\
&& Llama2 & Llama2 &\cellcolor{red!25} 0.655$\to$0.652 & 0.813$\to$0.818  \\  \cline{2-6} 
& \multirow{4}{*}{Llama2} & Gemini & Gemini & 0.641$\to$0.660 & 0.843$\to$0.863 \\
&& Llama2 & Gemini & 0.641$\to$0.649 & 0.843$\to$0.856  \\
&& Gemini & Llama2 & 0.641$\to$0.657 & 0.843$\to$0.869  \\
&& Llama2 & Llama2 & 0.640$\to$0.648 & 0.843$\to$0.856  \\ \midrule
\multirow{8}{*}{Gemini} & \multirow{3}{*}{GPT3.5} & GPT3.5 & GPT3.5 & 0.718$\to$0.721 & \cellcolor{red!25}0.850$\to$0.835  \\
&& Llama2 & GPT3.5 &\cellcolor{red!25} 0.714$\to$0.712 &\cellcolor{red!25} 0.836$\to$0.833  \\
&& GPT3.5 & Llama2 & 0.719$\to$0.721 & \cellcolor{red!25}0.840$\to$0.824  \\
&& Llama2 & Llama2 &\cellcolor{red!25} 0.717$\to$0.716 &\cellcolor{red!25} 0.840$\to$0.836  \\ \cline{2-6}
& \multirow{4}{*}{Llama2} & GPT3.5 & GPT3.5 & 0.558$\to$0.573 & 0.917$\to$0.919  \\
&& GPT3.5 & Llama2 & 0.592$\to$0.600 & 0.965$\to$0.965 \\
&& Llama2 & GPT3.5 & 0.592$\to$0.593 & \cellcolor{red!25}0.965$\to$0.959  \\
&& Llama2 & Llama2 &\cellcolor{red!25} 0.592$\to$0.590 & 0.965$\to$0.965  \\ \midrule
\multirow{8}{*}{Llama2} & \multirow{3}{*}{GPT3.5} & GPT3.5 & GPT3.5 & 0.606$\to$0.640 & 0.724$\to$0.743  \\
&& Gemini & Gemini & 0.604$\to$0.634 & 0.721$\to$0.736  \\
&& Gemini & GPT3.5 & 0.606$\to$0.628 & 0.724$\to$0.734  \\
&& GPT3.5 & Gemini & 0.601$\to$0.641 & 0.718$\to$0.763  \\ \cline{2-6}
& \multirow{4}{*}{Gemini} & GPT3.5 & GPT3.5 & 0.534$\to$0.569 & 0.714$\to$0.737  \\
&& Gemini & GPT3.5 & 0.558$\to$0.583 & 0.756$\to$0.761  \\
&& GPT3.5 & Gemini & 0.534$\to$0.575 & 0.729$\to$0.789  \\
&& Gemini & Gemini & 0.554$\to$0.599 & 0.747$\to$0.785  \\ \bottomrule
\end{tabular}%
}
\caption{Different LLMs to construct FIDES, which evaluated on Mintaka dataset, the red cell represents the abnormal editing. AG is a long form Answer Generation module.}
\label{Mintaka results}
\end{table}

\subsection{Analyzing the ability of LLMs to act in different modules}

\label{Analyzing the ability}

FIDES is a model-agnostic and versatile framework comprising four modules that can be adapted with any combination of LLMs. To assess the capabilities of different LLMs in tasks like Fact Decomposition (FD) and Factual Edit (FE), we conducted comprehensive experiments on three datasets (Mintaka, NQ, and Musique) using GPT-3.5, Gemini, and either Llama2 or Llama3. For example, in our configurations, we employed GPT-3.5 for generating long form answers, Gemini for FD, Llama3 70B for the EV module, and Gemini again for FE. By leveraging these combinations of LLMs, we varied the configurations of FIDES to enhance answer attribution capabilities. 

To illustrate our findings, we visualize the key information of the experiments (Due to page limitation, the partial experimental results are shown in Table \ref{Mintaka results}). As shown in Figure \ref{fig4}, GPT-3.5 consistently achieves the highest AF1 scores compared to Gemini, Llama2, and Llama3 in the original answers, except for the NQ dataset where Gemini performs better. Specifically, GPT-3.5's performance is only 0.006 points lower than Llama3's in this instance.

The knowledge boundary problem in LLMs often leads to hallucinations, which can result in abnormal editing. We further analyzed the proportions of abnormal editing by GPT-3.5-turbo, Gemini, Llama2, and Llama3, which are 43.75\%, 35\%, 35\%, and 20\%, respectively, when used in the FE module. This analysis reveals that GPT-3.5-turbo makes the most mistakes, whereas Llama3 70B exhibits the least abnormal editing behavior.

We aim to identify the best module for long form answer generation. GPT-3.5-turbo excels in KGQA and open-domain QA, while Gemini performs best in multi-hop QA according to the results of baseline DRA. As shown in Figure \ref{fig4}, GPT-3.5-turbo demonstrates the strongest generalization ability in the FD module, and Llama3 70B is the most stable model for editing. Utilizing the optimal settings of these LLMs (GPT-3.5-turbo as FD and Llama3 70B for EV and FE), we constructed FIDES. The experimental results are displayed in Table \ref{different_llm}  and Figure \ref{fig4}. The overall performance across the six datasets is superior to that achieved with a single LLM. 
\begin{table}[ht]
\centering
\caption{The experimental results on six datasets using multiple different LLMs to build our framework.}
\setlength{\tabcolsep}{0.63cm}{
\begin{tabular}{clll}
\toprule
\multicolumn{1}{l}{} & \multicolumn{1}{c}{AR} & \multicolumn{1}{c}{AP} & \multicolumn{1}{c}{AF1} \\ \midrule
WebQSP & 0.807 & 0.865 & 0.835 \\
Mintaka & 0.708 & 0.759 & 0.733 \\
NQ & 0.704 & 0.760 & 0.731 \\
StrategyQA & 0.638 & 0.702 & 0.668 \\
Hotpotqa & 0.627 & 0.660 & 0.643 \\
Musique & 0.537 & 0.567 & 0.552 \\ \bottomrule
\end{tabular}%
}

\label{different_llm}
\end{table}

\begin{figure}[ht!]
  \centering 
  \includegraphics[height=3.3cm,width=8.2cm]{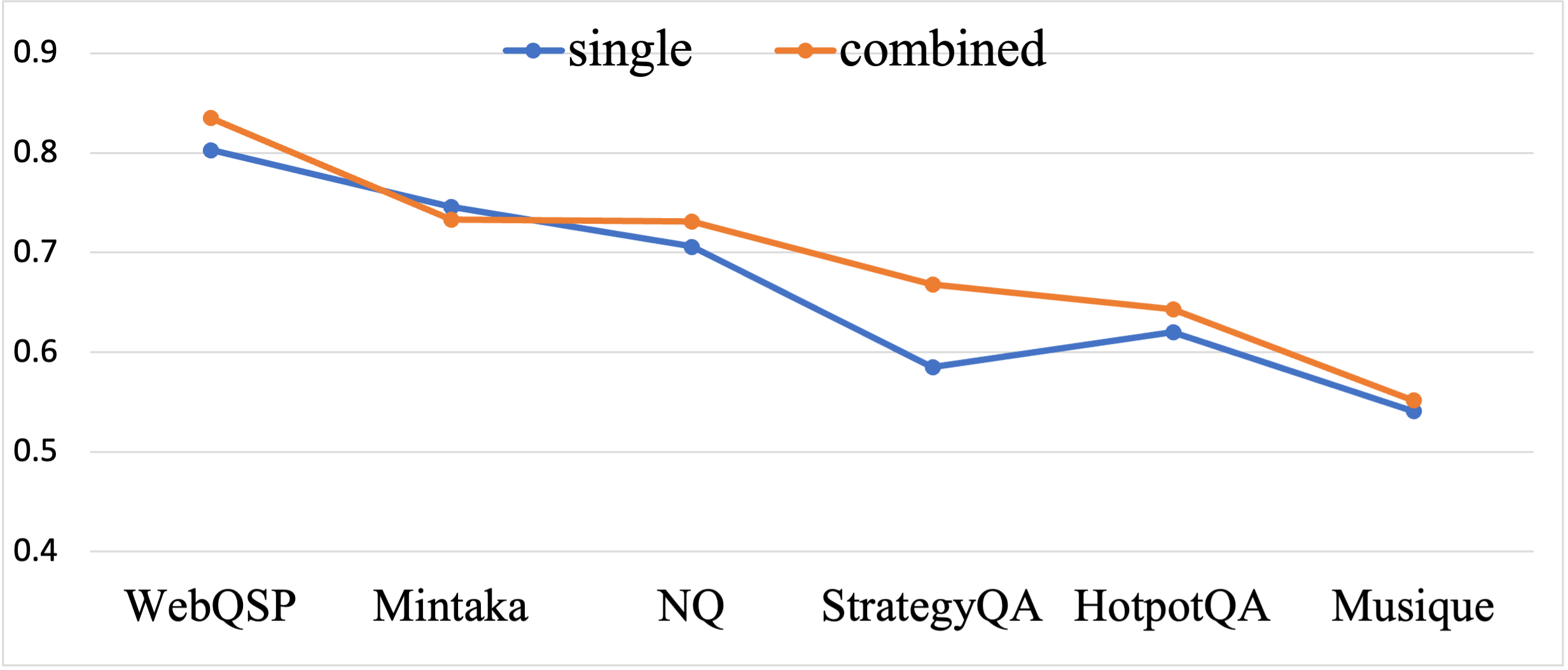}
  \caption{The results of FIDES using a single LLM vs combining different LLMs. The former results are obtained by averaging the performance across 
  multiple single LLMs. }
  \label{single_combined}
  \end{figure}

\subsection{Binary classification metric evaluation}
We also report binary classification evaluation results of accuracy with AutoAIS\cite{bohnet2022attributed} in Table \ref{AutoAIS}, which is a metric to evaluate the AIS score among the question, answer and evidence by NLI model\citep{honovich-etal-2022-true}, ``1'' for support and "0" for not support. For KGQA and Open-domain QA, post-hoc attributed framework achieved better performance than multi-hop QA datasets, the main reason for this phenomenon is that the multi-hop QA task is a challenging problem for LMs, but it is even more challenging from the perspective of post-attribution and is still a topic worth exploring in depth.

\begin{table}[]
\centering
\setlength{\tabcolsep}{0.23cm}{
\begin{tabular}{ccccc}
\toprule
\diagbox{Datasets}{Methods} & DRQA & DRA & RARR & FIDES \\ \midrule 
WebQSP & 0.553 & 0.546 & 0.513 & \textbf{0.713} \\
Mintaka & 0.387 & 0.380 & 0.407 & \textbf{0.500} \\
NQ & 0.420 & 0.340 & 0.349 & \textbf{0.500} \\
StrategyQA & 0.081 & 0.094 & 0.128 & \textbf{0.168} \\
HotpotQA & 0.033 & 0.060 & 0.167 & \textbf{0.200} \\ \bottomrule
\end{tabular}%
}
\caption{Binary classification results with GPT-3.5-turbo}
\label{AutoAIS}
\end{table}

\subsection{Human evaluation}
\label{sec:eval}

To assess the rationality of our proposed metric $Attr_{auto-P}$ and AR and AF1, we randomly select 50 samples in the Mintaka, NQ and Musique test dataset and let three annotators judge. We follow the process of AIS \citep{Rashkin2021MeasuringAI}, the binary classification is used to judge whether there is an entailment relationship between evidence and long form answer. The rules are as follows: For every evidence snippet, it should compare with every sentence in the answer; if it can not support total sentences, it will be defined as noise evidence. On the contrary, it will be designated as valid evidence. Furthermore, we also report the Pearson correlation coefficient between automated and human evaluations. From Table \ref{human_eval}, the result of the human evaluation is close to the automated assessment.
\begin{table}[htp]
\centering
\setlength{\tabcolsep}{0.5cm}{
\begin{tabular}{cccc}
\toprule
Evaluation metric             & Mintaka & NQ & Musique    \\ \midrule
AR    & 0.539   & 0.750 & 0.573 \\
AP    & 0.902   & 0.832 & 0.807 \\
AF1   & 0.856   & 0.824 & 0.725 \\ \bottomrule
\end{tabular}
}
\caption{\label{human_eval}
Pearson r-correlation of human evaluation results with AR, AP and AF1.}
\end{table}

\section{Case study}

\label{case_study}
\begin{figure}[h]
  \centering

  \includegraphics[height=5.2cm,width=8.2cm]{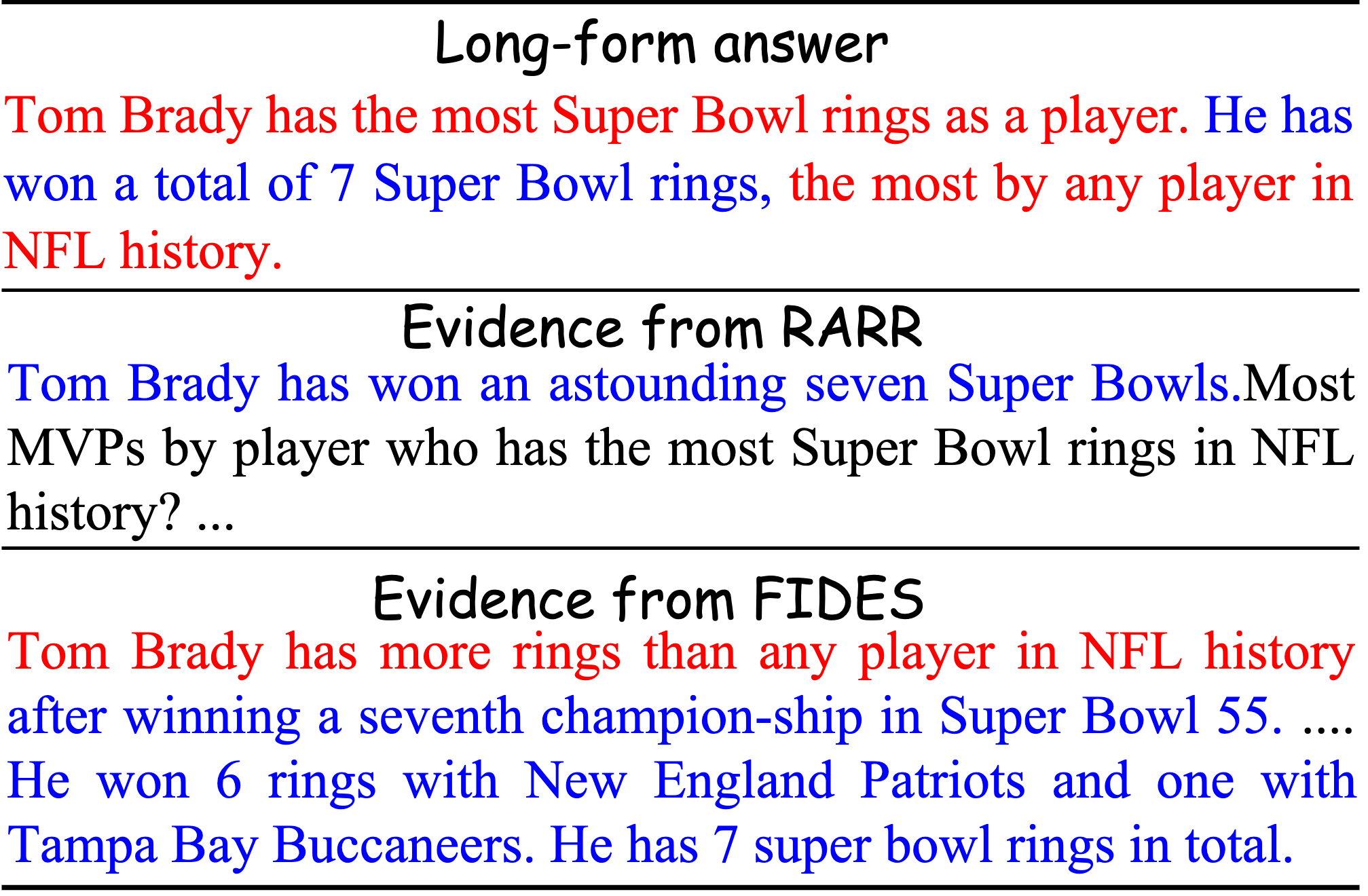}
  \caption{Retrieval result from RARR and FIDES.}
  \label{case3}
  \end{figure}

When comparing the retrieved evidence of RARR and FIDES in Figure \ref{case3}, FIDES tends to retrieve explicitly supportive content for the answer. In contrast, RARR may select relevant evidence that cannot be explicitly expressed. For instance, in supporting ``\textit{Tom Brady has the most Super Bowl rings as a player.}'' FIDES offers more direct evidence, whereas RARR provides evidence that necessitates additional reasoning. This presents increased demands and challenges for automatic evaluation models. 

\begin{figure}[h]
  \centering 
  \includegraphics[height=9.5cm,width=8.2cm]{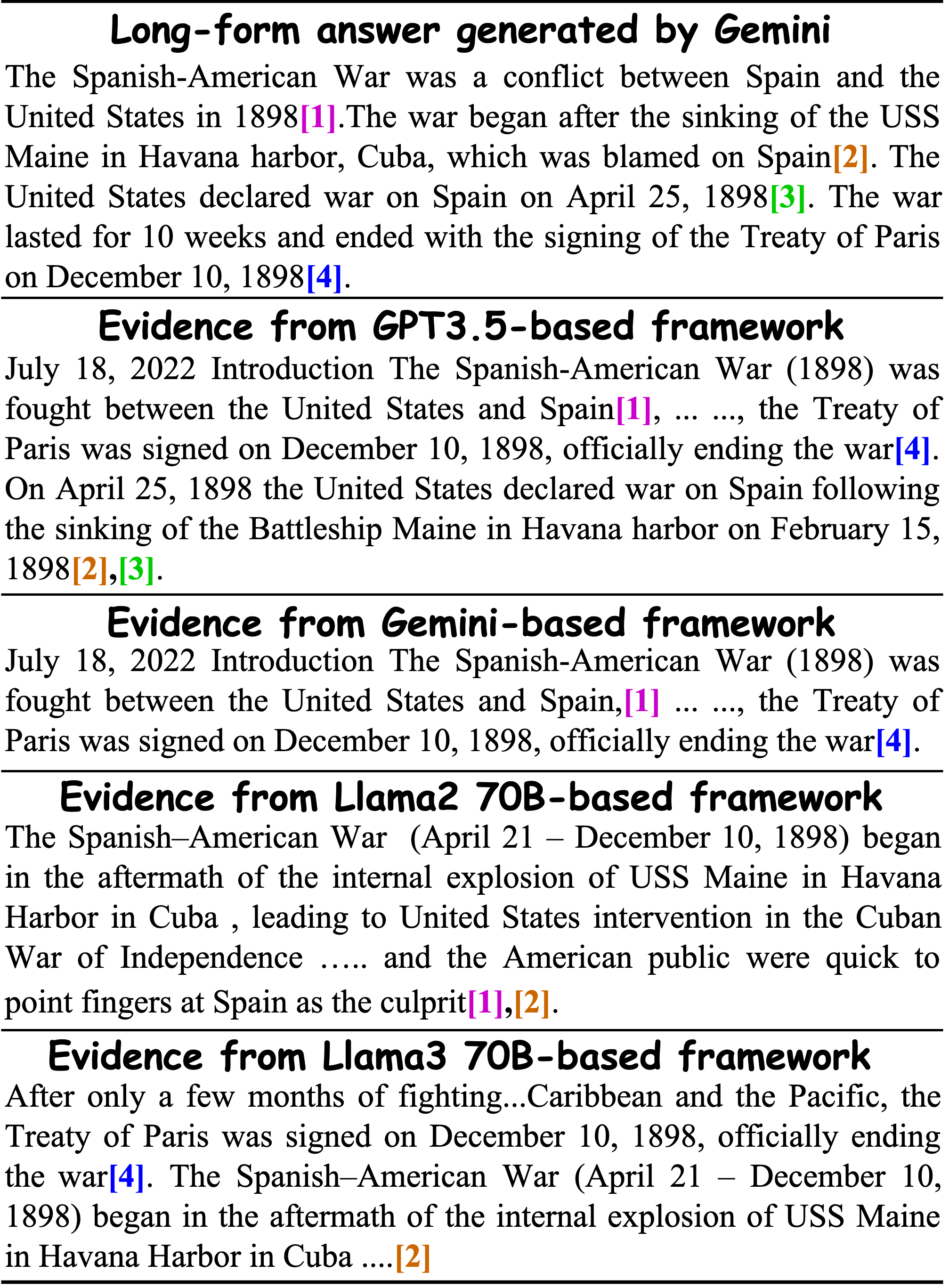}
  \caption{The evidence from different LLMs according to the answer generated by Gemini. Serial numbers in the same color represent sentence-evidence pairs.}
  \label{case4}
  \end{figure}

As shown in Figure \ref{case4}, we compare the four evidence snippets from systems based on GPT-3.5, Gemini, Llama2, and Llama3. For a long form answer generated by Gemini, which consists of four sentences, the GPT-3.5-based framework, after fact decomposition with SS and CD, retrieves the most relevant evidence snippets to cover all the facts in the answer. This demonstrates that GPT-3.5-turbo, when used as the SS and CD modules, outperforms the other LLMs.

\section{Conclusion}
We propose a novel post-hoc retrieval framework FIDES to attributed question answering with LLMs. The effectiveness of this framework is demonstrated across six datasets using 
four prominent LLMs. One key finding is that fact decomposition plays in important role in the post-hoc retrieval approach.  We also observe the performance differences of LLMs in different modules of FIDES. The best combination of LLMs for FIDES outperforms the single best LLM for FIDES significantly, in particular on some of the harder dataset (StrategyQA).  

\section*{Limitation}
The experimental results show that the \textbf{FE} module still has space for further improvement. Instruct-tuning of LLMs can be explored in the future. 
In addition, we have evaluated the performance of the answer generator, \textbf{SS} and \textbf{CD} and \textbf{FE}, but the module of \textbf{EV} can not be directly evaluated due to the lacking of labeled datasets. It is urgently required to develop more benchmarks or investigate non-labor-intensive evaluation methods to further improve the practicability of post-hoc attributed question answering. Using search engines to retrieve evidence may face irrelevant or partially support situations, which may be caused by sub-optimal retrieval. 




\begin{ack}
We thank the reviewers for their insightful comments. This research was supported by NSF China (No.62376144), by Science and Technology Cooperation and Exchange Special Project of Shanxi Province (No.202204041101016) , by Key Research and Development Project of Shanxi Province, by Natural Language Processing Innovation Team (Sanjin Talents) Project of Shanxi Province and by Fundamental Research Program of Shanxi Province (No.202403021211092).
\end{ack}


\bibliography{m4851}

\end{document}